\documentclass {article} 
\usepackage{nips16submit,times}
\usepackage{graphicx}
\graphicspath{ {Figures/} }
\usepackage{hyperref}
\usepackage{amsmath}
\usepackage{url}
\usepackage{wrapfig}
\usepackage{caption}
\usepackage{subcaption}
\usepackage{xcolor}

\title{Filling in the details: Perceiving from low fidelity images}

\author{
Farahnaz Ahmed Wick\thanks{ Email: fwick@cs.umb.edu} \\
University of Massachusetts Boston,\\ 
Boston, MA 02125 \\
\texttt{fwick@cs.umb.edu} \\
\And
Michael L. Wick \\
University of Massachusetts Amherst, \\
Amherst, MA 01003 \\
\texttt{mwick@cs.umass.edu} \\
\And
Marc Pomplun \\
University of Massachusetts Boston, \\
Boston, MA 02125 \\
\texttt{mpomplun@cs.umb.edu} \\
}

\newcommand{\cut}[1]{}
\newcommand{\hidden}{h}
\newcommand{\oimage}{x}
\newcommand{\fimage}{\hat{x}}
\newcommand{\gimage}{y}
\newcommand{\foveate}{\phi}
\newcommand{\autoencode}{f}
\newcommand{\encode}{f_e}
\newcommand{\decode}{f_d}
\newcommand{\rec}{\text{g}}
\newcommand{\loc}{\ell}
\newcommand{\saccade}{s}
\newcommand{\weights}{W}
\newcommand{\sigmoid}{\sigma}

\nipsfinalcopy 

\begin{document}

\maketitle

\begin{abstract}
  Humans perceive their surroundings in great detail even though most of our visual field is reduced to low-fidelity color-deprived (e.g. dichromatic) input by the retina. In contrast, most deep learning architectures are computationally wasteful in that they consider every part of the input when performing an image processing task.  Yet, the human visual system is able to perform visual reasoning despite having only a small fovea of high visual acuity. With this in mind, we wish to understand the extent to which connectionist architectures are able to learn from and reason with low acuity, distorted inputs. Specifically, we train autoencoders to generate full-detail images from low-detail ``foveations'' of those images and then measure their ability to reconstruct the full-detail images from the foveated versions. By varying the type of foveation, we can study how well the architectures can cope with various types of distortion. We find that the autoencoder compensates for lower detail by learning increasingly global feature functions. In many cases, the learnt features are suitable for reconstructing the original full-detail image. For example, we find that the networks accurately perceive color in the periphery, even when 75\% of the input is achromatic.

\end{abstract}

\section{Introduction}
The success of machine learning algorithms depends heavily upon the representation of the input data. A major appeal of deep learning, on which the current dominant approaches for machine vision tasks are based \cite{krizhevsky2012imagenet}, is that they can automatically learn useful feature representations from the data. A criticism of most deep architectures is that they wastefully process every input component when performing a task; for example, the input layer considers all pixels in every region of the input when learning an image classifier and making classification decisions. 

In contrast, the human visual system has only a small fovea of high resolution chromatic input allowing it to more judiciously budget computational resources \cite{lennie2003cost}. In order to receive additional information in the field of view, we make either covert or overt shifts of attention. Overt shifts of attention or \textit{eye-movements} allow us to bring the fovea over particular locations in the environment that are relevant to current behavior. To avoid the serial nature of processing as demanded from overt shifts of attention, our visual system can also engage in covert shifts of attention in which the eyes remain fixated on one location but attention is deployed to a different location.

The human retina receives 10 million bits per second which exceeds the computational resources available to our visual system to assimilate at any given time \cite{koch2006much}. Even though we perceive the environment around us in great detail, only a small fraction of the information registered by the visual system is processed. This paper asks a simple question: If high detail input were not available, would artificial neural networks still be able to capture aspects of the underlying distribution? 

\begin{figure}[h]

\begin{center}
\includegraphics[width= 0.8\textwidth]{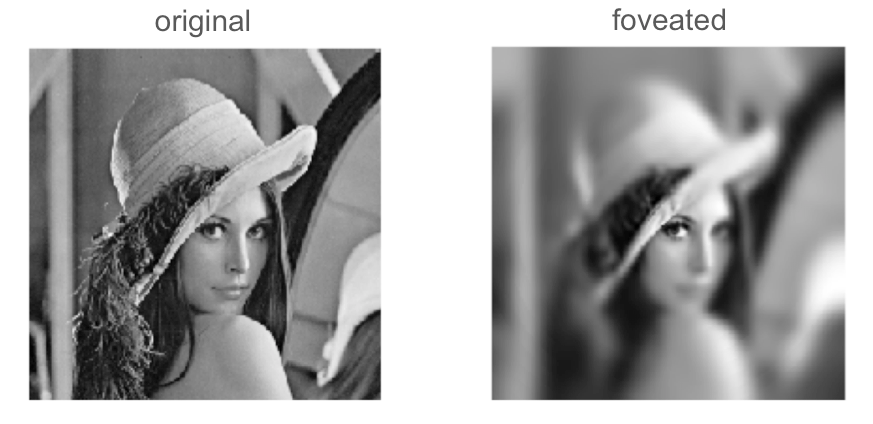}
\end{center}
\caption{Current deep networks are receiving images on the left as input whereas the human visual system receives foveated images as the one on the right.}
	\label{fig: foveated}
\end{figure}

To further put this question in perspective, our own fovea takes up only 4\% of the entire retina \cite{michels1990retinal} and is solely responsible for sharp central full color vision with maximum acuity. The relative visual acuity diminishes rapidly with eccentricity from the fovea \cite{cowey1974human}. As a result, visual performance is best at the fovea and progressively worse towards the periphery \cite{low1946some}. Indeed, our visual cortex is receiving distorted color-deprived visual input except for the central two degrees of the visual field \cite{hansen2009color} as seen in Figure \ref{fig: foveated}. Despite receiving such a distorted signal, we perceive the world in color and high resolution and are mostly unaware of this distortion. Even when confronted with actual blurry or distorted visual input, our visual system is good at extracting the scene contents and context. For instance, our system can recognize faces and emotions expressed by those faces in resolutions as low as 16 x 16 pixels \cite{sinha2006face}. We can reliably extract contents of a scene from the gist of an image \cite{oliva2005gist} even at low resolutions \cite{potter1969recognition,judd2011fixations}. 

Recently, Ullman et al. \cite{ullman2016MIRC} has shown that our visual system is capable of recognizing contents of images from critical feature configurations (called minimal recognizable images or MIRCs) that current deep learning systems cannot utilize for similar tasks. These MIRCS resemble foveations on an image and their results reveal that the human visual system employs features and processes that are not used by current deep networks. Similarly, little attention has been given by the deep learning community to how these networks deal with distorted or noisy inputs. We draw inspiration from the abilities of the human visual system and ask whether an artificial neural network can learn to perceive an image from low fidelity input. If this is the case, we can design state of the art architectures in image super resolution, automatic image coloring, image compression and at the same time, reduce computational costs of processing entire images associated with deep networks.

There has been a revival in applying the idea of attention to deep learning architectures \cite{larochelle2010learning,mnih14recurrent,bahdanau14neural,xu15show}. Such work is exciting and has lead to improvements in tasks ranging from machine translation \cite{bahdanau14neural} to image captioning \cite{xu15show}. However, in many approaches---especially those that employ a \textit{soft} attention mechanism---the computational cost is increased. For example, when generating a target sentence, a network must compute a softmax over every word of a source sentence or location of a source image. Unlike these systems, humans perceive by sequentially directing attention to relevant portions of the data and in turn enables our visual system to reduce computational costs \cite{renninger2007look,koch2006much}.

In this paper, we want to understand what kind of information can be gleaned from low-fidelity inputs. What can be gleaned from a single foveal glimpse? What is the most predictive region of an image? We present a framework for studying such questions based on a generative model known as an autoencoder. In contrast to traditional or de-noising autoencoders \cite{vincent08extracting}, which attempt to reconstruct the original image (or respectively, a salt and pepper corrupted version), our autoencoders attempt to reconstruct original high-detail inputs from lower-detail foveated versions of those images (that is, images that are entirely low detail except perhaps a small ``fovea'' of high detail). Thus, we have taken to calling them defoveating autoencoders (DFAE). 

We find that even relatively simple DFAE architectures are able to perceive color, shape and contrast information, but fail to recover high-frequency information (e.g., textures) when confronted with extremely impoverished input. Interestingly, as the amount of detail present in the input diminishes, the structure of the learnt features becomes increasingly global.

\section{Background: Autoencoders}

Autoencoders are a class of unsupervised algorithms which pairs a bottom-up recognition network (encoder) with a top-down generative network (decoder). The encoder, denoted as the function $f_{\theta}$, forms a compressed representation $\mathbf{y} = f_{\theta}(\mathbf{x})$ of the input $\mathbf{x}$. $\mathbf{y}$ is the feature vector representation or code computed from $\mathbf{x}$. In the context of our work, we were interested in whether or not we can learn a rich representation $\mathbf{y}$ of a low fidelity input image $\mathbf{x}$.

The output, denoted as the function $g_{\theta'}$, maps the feature vector back into the input space producing a reconstruction $\mathbf{z} = g_{\theta'}(\mathbf{y})$ through the minimization of a reconstruction error function. Good generalization means reconstruction error of test examples should be close to the reconstruction error for training examples. To capture the structure of the underlying data distribution and prevent the autoencoder from learning the identity function, we can either require the hidden layer to have lower dimensionality than the input or regularize the weights \cite{vincent2010stacked}. The lower dimension constraint is what the classical autoencoder or PCA does while the higher dimension is used by the sparse autoencoders \cite{poultney2006efficient}. Recently, denoising autoencoders have been shown to regularize the network by adding noise such as salt-and-pepper (SP) noise to the input, thus forcing the model to learn to predict the true image from its noisy representation \cite{vincent2010stacked}.

In summary, the basic autoencoder training consists of optimizing parameter vector $\theta$ to minimize reconstruction error as measured by the loss, $L$:
\begin{equation}
\label{eq:1}
\sum_{i}{ L(x^{i},g_{\theta'}(f_{\theta}(x^{i})))}
\end{equation}
where $x^i$ is a training example. The minimization is carried out by standard gradient descent algorithms like backpropagation. The commonly used forms for the encoder is an affine mapping followed by non linearity:
\begin{equation}
\label{eq:2}
f_{\theta}(\mathbf{x}) = s(\mathbf{Wx} + \mathbf{b})
\end{equation}

where  $s$ is the encoder activation function, $\theta = \{\mathbf{W,b}\}$, $\mathbf{W}$ is the weight matrix and $\mathbf{b}$ is the bias vector. Similarly the decoder mapping is: 

\begin{equation}
\label{eq:3}
g_{\theta'}(\mathbf{y}) = s(\mathbf{W'y} + \mathbf{b'})
\end{equation}

with the appropriately sized parameters $\theta' = \{\mathbf{W'},\mathbf{b'}\}$

As mentioned above, it has been shown that the features learnt by the encoder without any non-linearity are a subspace of the principal components of the input space \cite{baldi1989neural}. However, when a non-linear activation such as a sigmoid is used in the encoder, an AE can learn more powerful feature detectors than a simple PCA \cite{japkowicz2000nonlinear}. The architecture of a simple one hidden layer AE is very similar to that of a multilayer perceptron (MLP). The difference between AEs and MLPs lies in the output layer: the MLP predicts the class $C$ of the input $X$ whereas an AE reconstructs $Z$ from $X$. 

We will start by reviewing related work on using distorted inputs to train deep networks and then move on to describe the architecture of AE that was used to test feature extraction from downsampled images.

\section{Related work}

Using noisy or \textit{jittered} inputs to understand feature learning in the framework of AEs or MLPs has been explored before \cite{lecun1989backpropagation,vincent2010stacked}. Vincent et al. \cite{vincent2010stacked} first proposed training autoencoders with corrupted image as input. Therefore their \textit{denoising} autoencoder (DAE) learnt to reconstruct the clean input from a corrupted version. They have shown that introducing noise to the input lowers classification error on benchmark classification problems. The filters produced by denoising AEs tend to capture more distinctive blob-like features and with higher level of corruption in the input image, they learn less localized filters. In fact, Bishop \cite{bishop1995training} has argued that in a linear system training with noise has a similar effect as training with a regularizer, such as an L2 weight decay. Another proposal to make autoencoders noise invariant is by Rifal et al. \cite{rifai2011contractive}. They improved on DAEs by adding a penalty term, called the contraction ratio, to the learnt mapping which makes the features learnt more robust and invariant to change of raw input. In the spirit of denoising AEs, we incorporate a form of noise in our input image. However, unlike the SP noise, our noise is generated from using a foveation function (described below) on the image. We investigated whether foveations acted as a strong regularizer for the AE like the SP noise, thus allowing us to use it in deep architectures. 

Denoising images has been investigated using architectures other than autoencoders. Xie et al. \cite{xie2012image} presented an approach to remove noise from corrupted inputs using sparse coding and deep networks pre-trained with DAEs. Their end to end system could automatically remove complex patterns like text from an image in addition to simple patterns like pixels missing at random.  The type of noise additions they investigated were white Gaussian noise, SP noise (flipping pixels randomly), and image background changes. Along the same lines, post deblurring denoising \cite{schuler2013machine} and using convolutional neural networks for natural image denoising of patterns such as specks, dirt and rain has been investigated \cite{jain2009natural}.

As mentioned above, low resolution images can be considered as a type of noisy input. In the domain of image super resolution, Cui et al. \cite{cui2014deep} used low resolution images interpolated to the size of the output image and AEs in their pipeline to restore resolution of these images. Their cascade model is not trained end-to-end and requires optimization of each layer individually. A similar approach by Dong et al. \cite{dong2014learning} improves on Cui et al.\cite{cui2014deep} by using convolutional neural networks and with an end-to-end system. Behnke et al. \cite{behnke2001learning} demonstrated that difficult non-linear image reconstruction from low resolution inputs can be learnt by hiearchical recurrent networks. From a given 28 x 28 handwritten digit image as input, their system can iteratively increase it's resolution to 64 x 64 output. 

Our work can be viewed as an image super resolution problem, in that our network learns mapping between low resolution and high resolution images. In contrast to existing approaches, our network is end-to-end differentiable and thus learns features automatically via backpropagation. Current approaches require manual engineering of features and image pre-processing on top of interpolations. Finally, we emphasize here that our goal is \textit{not} achieving state of the art results in image super resolution. Rather, we want to study a deep architecture's ability to extract useful representation from low-detail images and showing the range in which mapping between low resolution and high resolution images is possible. The usefulness of the representation is then measured using mean squared error between input and reconstructed output.

\section{Framework: Defoveating Autoencoders (DFAE)}
We now present a framework for studying the extent to which neural networks can ``perceive'' an image given various types of low-detail (or foveated) inputs. We begin by specifying a space of neural network architectures and by precisely defining a notion of {\em perceives} that we can measure. It is important that the framework is general and not dependent on a specific task such as image classification in which, for example, the ability to learn domain-specific discriminating features might make it easy to solve the classification problem without fully modeling the structure of the input. This is undesirable because then we are unable to trust classification accuracy as a reliable surrogate for {\em perceiving}.

With this in mind, we focus instead on generative models of the raw input data itself, specifically autoencoders (AE). The AE's hidden units $\hidden$ are analogous to the intermediate neurons in our visual system that capture features and structure of the visual input. Similarly, the AE's weights $\weights$ forge visual memories of the training set and are thus analogous to long-term memory.  When these weights are properly trained, the activations of the hidden units reflect how the network is perceiving a novel input. However, since these units are not directly interpretable, we indirectly measure how well the network perceives by evaluating the similarity between the original and generated (high-detail) images: the more similar the images are, the better the network is able to perceive.

More formally, let $\oimage$ be the original input image and $\fimage=\foveate(\oimage)$ be a lower-detail {\em foveated} version of that image. That is, a version of the image which is mostly low-detail (e.g., downsampled, black-and-white, or both) except for possibly a small portion which is high-detail (mimicking our own fovea). For example, if we encode images as vectors of floats between 0 and 1 (reflecting pixel intensities in RGB or grayscale) then we might define a class of foveation functions as $\foveate:[0,1]^n\rightarrow [0,1]^m$ s.t. $m\ll n$ and the foveation function might downsample the original image according to the eccentricity from the image center while also removing most of the vector components corresponding to color. We then employ the autoencoder to defoveate $\fimage$ by generating a high-quality output image $\gimage = \autoencode(\fimage;\weights)$ in which, for example, $\gimage\in [0,1]^n$. Finally, we can then measure the similarity between $\gimage$ and $\oimage$ as:
\begin{enumerate}
\item a surrogate for how well the network perceives from the foveated input and 
\item part of a loss function to train the network.  
\end{enumerate}

In summary, DFAEs simply comprise:
\begin{enumerate}
\item A foveation function that filters an original image by removing detail (color, downsampling, blurring, etc). We will later make this the independent variables in our experiments so we can study the effect of different types of input distortion.
\item An autoencoder network that inputs the low-detail foveated input, but is trained to output the high-detail original image.
\item A loss function for measuring the quality of the reconstruction against the original image and for training the network.
\end{enumerate} 

Given this framework we can now study how well different architectures are able to cope with different types of foveated inputs. Note that much like denoising autoencoders, these autoencoders reconstruct an original image from a corrupted version. However, the form of corruption is a systematic foveation instead of random SP noise. Thus, as an homage to denoising autoencoders \cite{vincent08extracting}, we have termed these models {\em defoveating} autoencoders or DFAEs. We describe our exact model in the next section.

\subsection{DFAE Architecture and Loss Function}
In our experiments, we study DFAEs with fully connected layers. That is, DFAEs of the form:
\begin{align}
\fimage &= \foveate (x) \quad (\text{no learnable parameters}) \\
\hidden^{(0)} &= \tanh\left(W^{(0)}\fimage\right) \\
\hidden^{(i)} &=\tanh\left(W^{(i)}\hidden^{(i-1)}\right) \quad \text{for} \, i = 1,\cdots,k-1\\
\gimage &= \sigmoid\left(W^{(k)}\hidden^{(k-1)}\right)
\end{align}

\noindent where $\sigmoid$ is the logistic function. The sigmoid in the final layer conveniently allows us to compare the pixel intensities between the generated image $\gimage$ and the original image $\oimage$ directly, without having to post-process the output values. We experiment with the number of hidden units per-layer as well as the number of layers.  For training, we could employ the traditional mean-squared (MSE) error or cross-entropy loss, but we found that the domain-specific loss function of peak signal-to-noise ratio (PSNR) yielded much better training behavior. The PSNR between generated image a $\gimage=\autoencode(\foveate(\oimage))$ and its original input $\oimage$ is defined as follows:
\begin{equation}
\label{eq:psnr}
L_{H}(\oimage,\gimage) = \log_{10}{\left(\frac{1}{\sqrt{MSE(\oimage,\gimage)}}\right)} \quad \text{where} \; MSE(\oimage,\gimage) = n^{-1}\oimage^T\gimage
\end{equation}

Network parameters were initialized at random in the range [-0.1,0.1] and loss was minimized by stochastic gradient descent with adagrad updates \cite{duchi2011adaptive}. Adaptive gradient descent, or adagrad, is a form of stochastic gradient descent that determines the per-feature learning rate dynamically during training. Adagrad calculates a different learning rate for each feature, allowing it to efficiently learn the weights even for features that rarely occur in the training data. The learning rate was initialized at 1.0 and was adjusted by adagrad during training. We performed 1000 epochs of training in all experiments.

\begin{figure}[!tbp]
  \begin{subfigure}[b]{0.54\textwidth}
    \includegraphics[width=\textwidth]{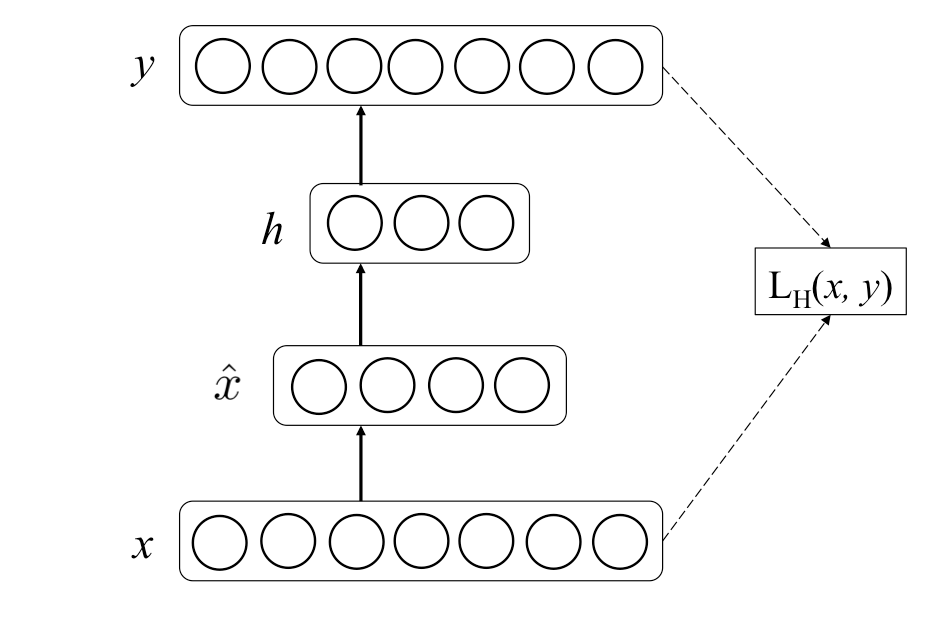}
    \caption{DFAE architecture}
    \label{fig: architecture}
  \end{subfigure}
  \hfill
  \begin{subfigure}[b]{0.45\textwidth}
    \includegraphics[width=\textwidth]{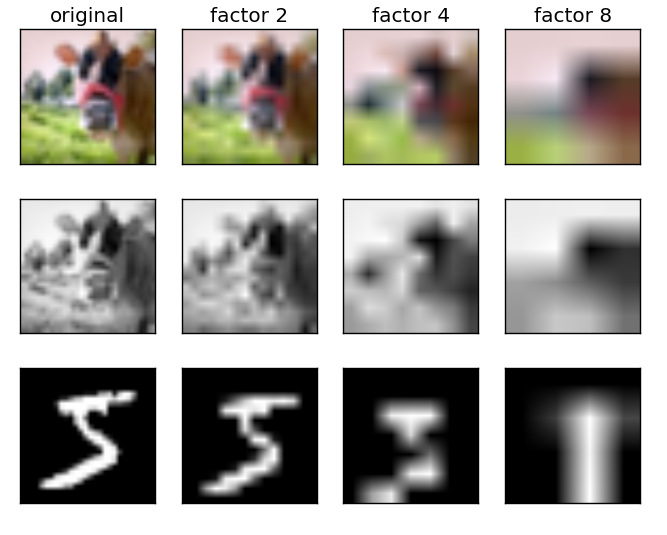}
    \caption{Input with no foveations}
    \label{fig: zeroFoveatedImages}
  \end{subfigure}
\caption{(a) An example of a fully-connected defoveating autoencoder architecture with a single hidden layer. An image $\oimage$ is foveated (via $\foveate(\oimage)$) to $\fimage$. The autoencoder then mapped $\fimage$ to hidden representation $\hidden$ from which the final output image $\gimage$ is generated $\gimage=\autoencode(\fimage;\weights)$. Reconstruction error is measured: $L_{H}(\oimage,\gimage)$. (b) Inputs with no foveations.}

\end{figure}

\subsection{Recurrent DFAEs for Sequences of Foveations}
The above architecture is useful for studying single foveations, which is the primary focus of this work. However, we remark that it is straightforward to augment DFAEs with recurrent connections to handle a sequence of foveations similar to what has has been done for solving classification tasks with attention \cite{mnih14recurrent}. First, augment the foveation function $\foveate$ to include a locus $\ell$ on which the fovea is centered. Second, a saccade function $\saccade(\hidden_t;\weights_{\saccade})$ predicts such a locus from the DFAE's current hidden states $\hidden_t$, and finally we make the hidden state recurrent via a function $\rec(\hidden_{t-1},\weights_{\rec})$. Putting this all together yields the following architecture:
\begin{align*}
\fimage_t &= \foveate(\oimage_t, \loc_t) & \color{gray}{\text{foveate the image at location } \ell} \\
\hidden_t &= \encode(\rec(\hidden_{t-1};\weights_{\rec}), \fimage_t;\weights) & \color{gray}{\text{encode: compute new hidden states}}\\
\ell_t &= \saccade(\hidden_t; \weights_{\saccade}) & \color{gray}{\text{compute new locus for next foveation}} \\
\gimage_t & = \decode(h_t) & \color{gray}{\text{decode: reconstruct high detail image}}
\end{align*}
Now the DFAE can handle a sequence of foveations for a given input image, further allowing us to train the model in a more realistic fashion. That is, the human visual system does not have access to all the high detail information at once and must must instead forge visual memories from a sequence of foveations. Thus, to mimick this, rather than trying to reconstruct the original image, we can instead try to reconstruct the foveation at time $t$ from the information available at time $t-1$. This is similar to how a language model is trained.

For now, we focus on studying the effects of single foveations using the non-recurrent form of the DFAE.

\section{Experiments}

Recall that we are interested in the question of whether an artificial neural network can \textit{perceive} an image from foveated input images. In the context of autoencoders, the hidden layers $\hidden$ are responsible for representing the foveated inputs $\oimage$. If the network learns a reasonable representation, then it should be able to produce a higher resolution output $\gimage$. We can then measure how similar the output of the network is to the original image to evaluate how well the network can \textit{perceive}. In these experiments, we fix the architecture of our network to the family described in the previous section and vary the type of foveation, the number of hidden units and the number of layers and study the learnt features and reconstruction accuracy. We address the following questions:

\begin{enumerate}
\item \label{q:perceive-details} Can the network perceive aspects of the image that are not present in the input? What can it perceive and  under what conditions?
\item \label{q:perceive-color} Can the network perceive color that is not present in the input? Does it need a small fovea of color to do so?
\item \label{q:capacity}  How much capacity is required to perceive missing details?
\item \label{q:features} How does the network compensate for foveated inputs? Does the foveation affect the learnt features?
\end{enumerate}

To answer these questions, we define several foveation functions as described in the following section.

\subsection{Foveation Functions}
In our experiments, we study several different foveation functions (described in more detail in the appropriate sections). In many cases, downampling is employed as part of the foveation function for which we employ the nearest neighbor interpolation algorithm. Nearest neighbor interpolation is a simple sampling algorithm which selects the value of the nearest point and does not consider the values of the neighboring points at all. As an interpolation algorithm, it generates poor quality or blocky images as there is no smoothing function. We picked nearest neighbor as our downsampling algorithm to test the worst case possible downsampled inputs on our system. Foveation functions include:

\begin{itemize}
\item {\bf downsampled factor $d$ (no fovea):} no fovea is present, the entire image is uniformly downsampled by a factor of $d$ using the nearest neighbor interopolation method. For example, a factor of $4$ transforms a 28x28 image ito a 7x7 image and approximatley 94\% of the pixels are removed. Note, in the case of color images, each channel (RGB) is separately downsampled resulting in color distortion.  The downsampling factors tested for MNIST were 2, 4 and 7, and for CIFAR100 dataset were 2, 4 and 8. See~\ref{fig: zeroFoveatedImages} for examples.
\item {\bf scotoma $r$ (SCT-R):} entire regions ($r = $ 25\%, 50\% and 75\%) of the image are removed (by setting the intensities to 0) to create a blind spot/region, but the rest of the image remains at the original resolution. We experiment with the location of the scotoma (centered or not).
\item {\bf fovea $r$ (FOV-R): } only a small fovea $r$ of high resolution ($r = $ 25\% or 6\%); the rest of the image is downsampled by a factor of $4$. Note that the special case of $r = 0\%$ is equivalent to downsampling the entire image uniformly.
\item {\bf achromatic $r$ (ACH-R): } only a region of size $r$ has color; color is removed from the periphery by averaging the RGB channels into a single grayscale channel.
\item {\bf fovea-achromatic $r$ (FOVA-R): } combines the fovea $r$ with the achromatic region: only the foveated region is in color, the rest of the image is in grayscale and downsampled by a factor of $4$. 
\end{itemize}

\subsection{Datasets and pre-processing}

We used two datasets in our experiments: MNIST and CIFAR100. The MNIST database consists of 28 x 28 handwritten digits and has a training set of 60,000 examples and a test set of 10,000 examples. Therefore each class has 6000 examples. The CIFAR100 dataset consists of 32 x 32 color images of 100 classes. Some examples of classes are: flowers, large natural outdoor scenes, insects, people, vehicles etc. Each class has 600 examples. The training set consists of 50,000 images and the test set consists of 10,000 images.

We trained DFAEs on the MNIST and CIFAR100 dataset (in grayscale and color). We normalized the datasets so that the pixel values are between 0 and 1 and additionally, zero-centered them. This step corresponds to local brightness and contrast normalization. Aside from this step, no other preprocessing such as patch extraction or whitening was applied.

\subsection{Baseline: Comparison with standard interpolation functions} 

\begin{figure}[h]
\begin{center}
\includegraphics[width= 0.98\textwidth]{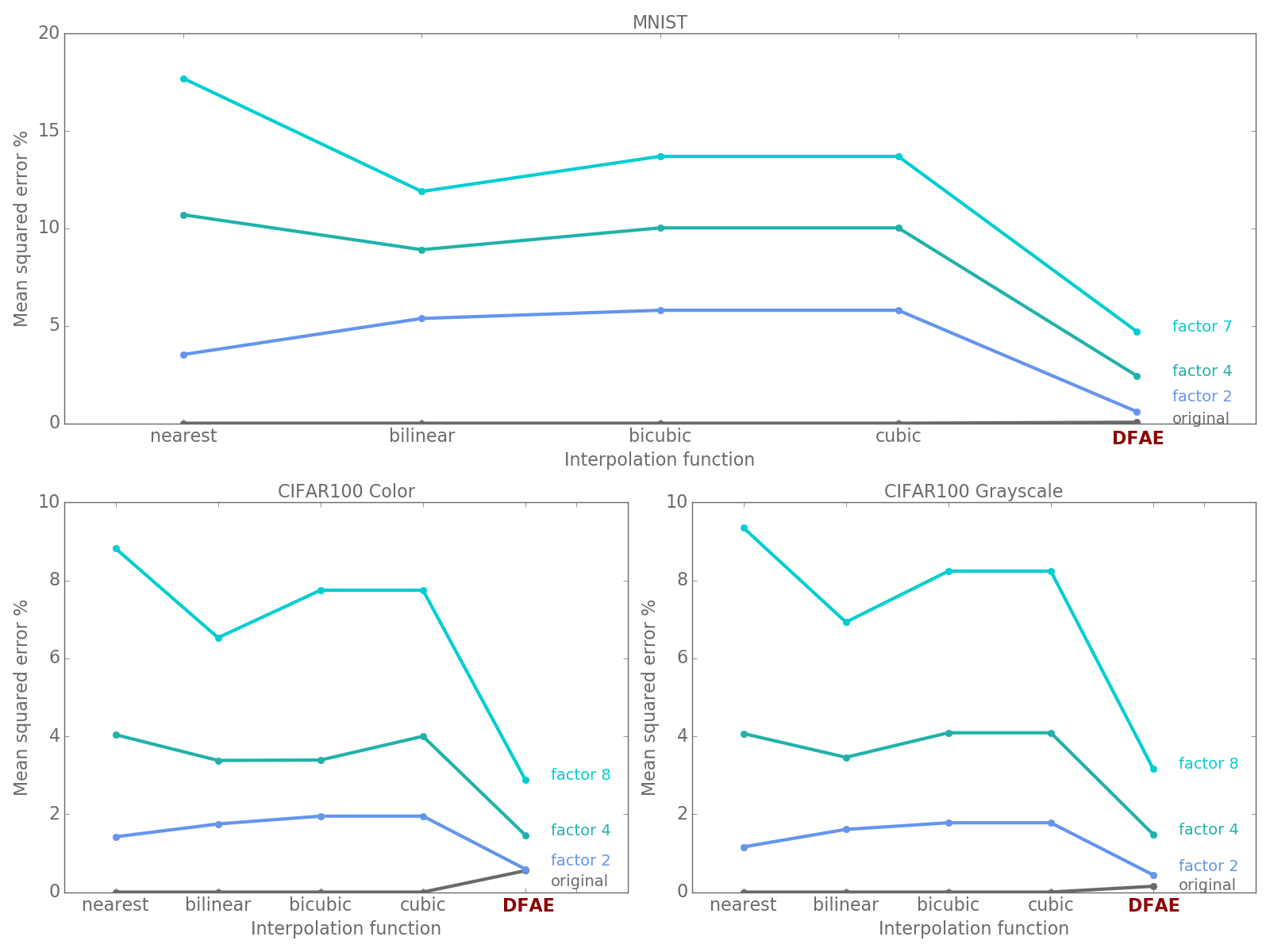}
\end{center}
\caption{Performance of standard upsampling algorithms compared to a single layer DFAE with 800, 1100, 3100 hidden units for MNIST, CIFAR100 grayscale and CIFAR100 color input respectively.}
	\label{fig: BaselinePlots}
\end{figure}

First, to establish baselines and context for our results, we compare a 1-layer DFAE to common upsampling algorithms found in image editing software. We report mean squared error (MSE) between the reconstructed image and original image to measure the quality of reconstructed images by the interpolation algorithm and the DFAE. An MSE of zero means the algorithm or DFAE is able to reconstruct the input with perfect accuracy. Figure \ref{fig: BaselinePlots} shows the MSE of the interpolation algorithms and a DFAE. Not surprisingly, the nearest neighbor performed the worst reconstruction overall. The bilinear interpolation performed the best in comparison to other upsampling algorithms tested. The interpolation algorithms performed poorly when they reconstructed an image that was downsampled beyond a factor of 2. The error rates produced by these interpolation algorithms on the MNIST dataset is higher than the natual image dataset. Figure \ref{fig: BaselinePlots} show that a single layer DFAE outperforms these standard algorithms for the datasets tested.

\subsection{Feature detectors learnt without any foveations}

Here we experiment with foveation functions in which the size of the fovea is 0; that is, these foveation functions uniformly downsample the original by various factors (the factors are experimentally controlled). The purpose of this experiment is to study how well the network can reconstruct the image when no high-detail input is available. The variables to consider are the number of hidden units per layer and the number of layers. Pilot experiments showed when the number of hidden units was less than the downsampled input size, DFAEs performed very poorly. This is not surprising because autoencoders cannot learn features in an \textit{under complete} state and the downsampled input contains impoverished features.

\begin{figure}[!tbp]
  \begin{subfigure}[b]{0.48\textwidth}
    \includegraphics[width=\textwidth]{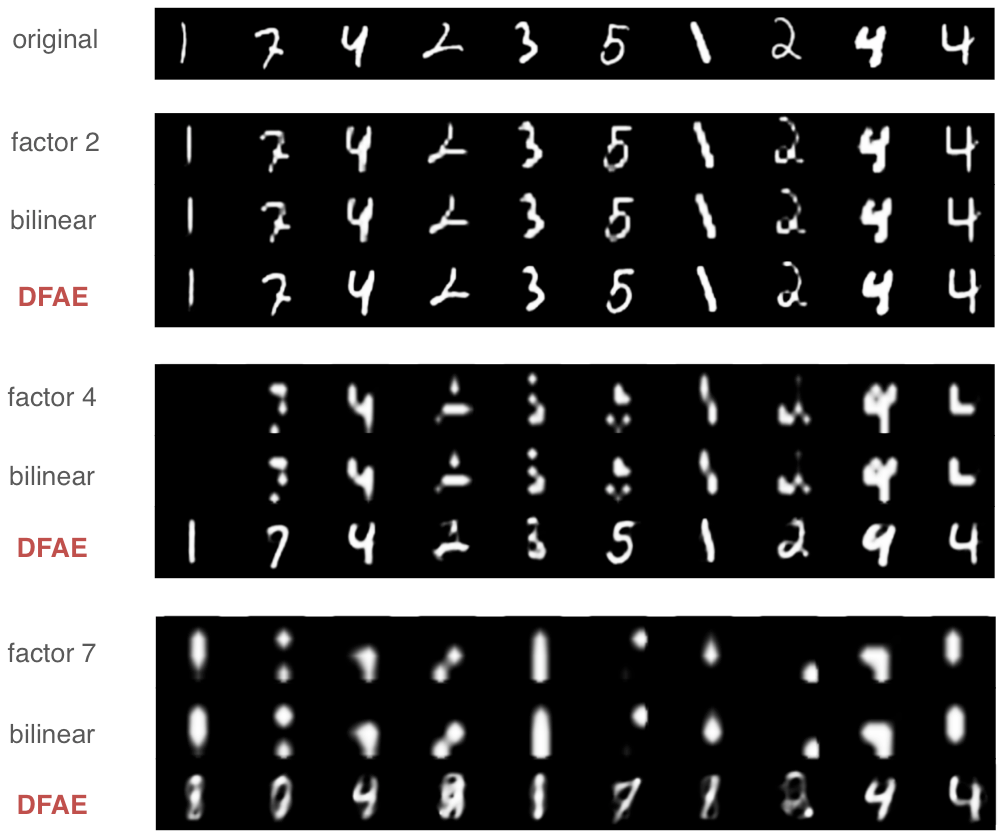}
    \caption{MNIST reconstruction}
    \label{fig: MNIST_output}
  \end{subfigure}
  \hfill
  \begin{subfigure}[b]{0.51\textwidth}
    \includegraphics[width=\textwidth]{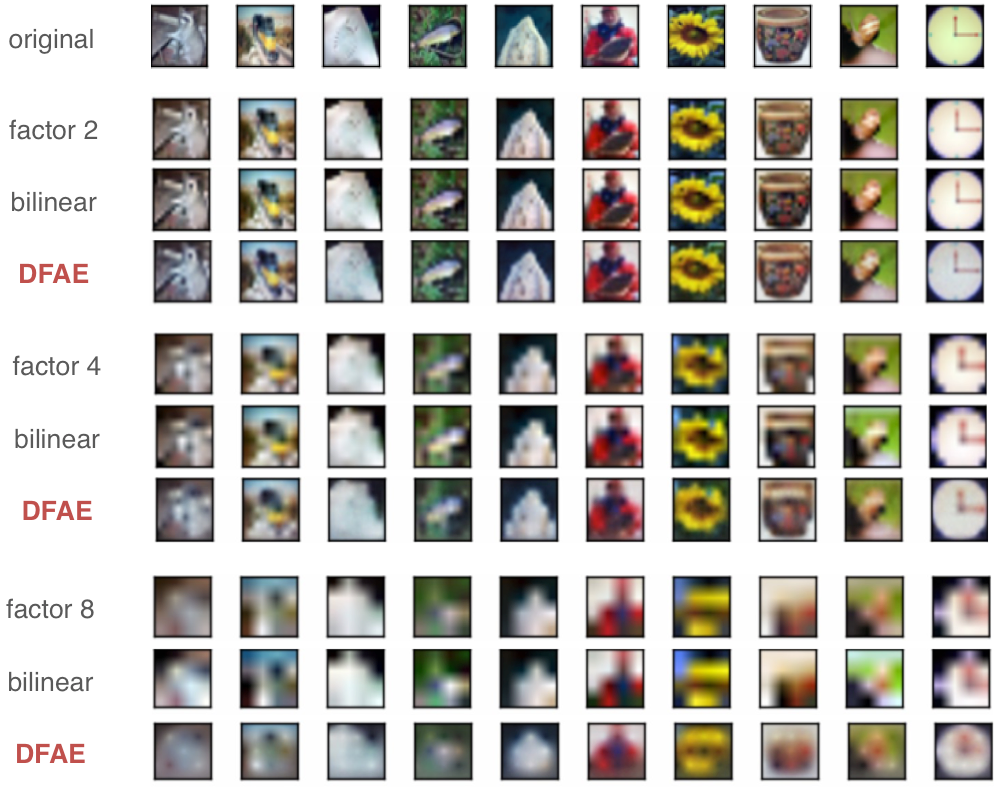}
    \caption{CIFAR100 reconstruction}
    \label{fig: CIFAR100_color_output}
  \end{subfigure}
  \caption{MNIST and CIFAR100 images reconstruction by DFAE. The top row shows the original image. Each row shows the downsampled input that was used during training, followed by reconstruction images by the bilinear algorithm and DFAEs.}
\label{fig: DownsampledReconstructions}
\end{figure}

Figure \ref{fig: DownsampledReconstructions} show examples of the reconstructed images by a single layer DFAE. The images produced by the DFAE is compared to upsampled reconstructions by the bilinear algorithm. When compared to the bilinear algorithm, DFAEs can correctly extract the contents of a downsampled input even when 94\% of pixels are removed. A compelling example is that even when faced with a blank input as seen in Figure \ref{fig: MNIST_output} the DFAE can correctly predict the digit $1$. However the performance of DFAEs suffered when the input was downsampled beyond factor $4$. Even though the DFAE made predictions based on the input, most of the reconstructions were incorrect. The reconstructed natural images as seen in Figure \ref{fig: CIFAR100_color_output} show that the DFAE learnt a smoothing and centering function even though it was unable to reconstruct the high frequencies in the images. The DFAEs could predict the shape of objects in the natural images but not the high frequency details.

\begin{figure}[!tbp]
  \begin{subfigure}[b]{\textwidth} 
  	\centering
    \includegraphics[width=0.8\textwidth]{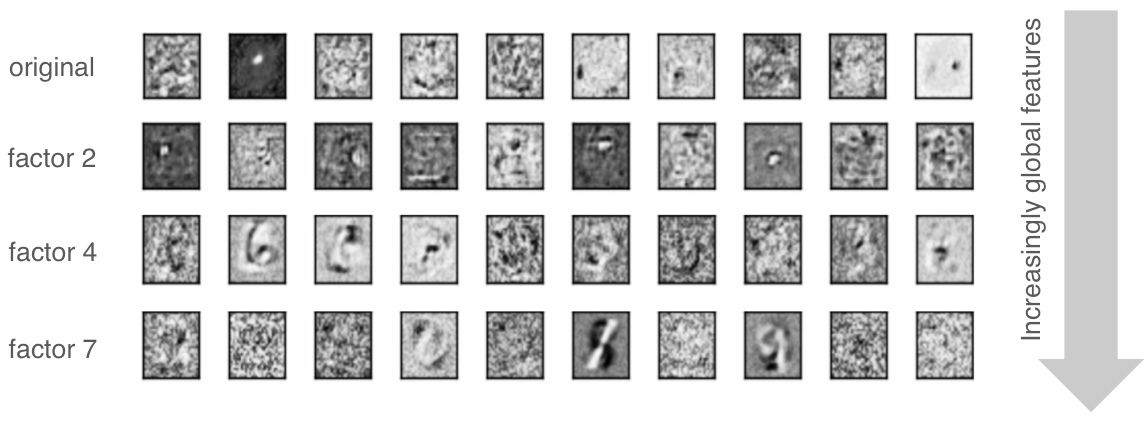}
    \caption{MNIST: 1-layer DFAE with 800 hidden units}
    \label{fig: MNIST_features}
  \end{subfigure}
  \hfill
  \begin{subfigure}[b]{\textwidth}
  	\centering
    \includegraphics[width=0.7\textwidth]{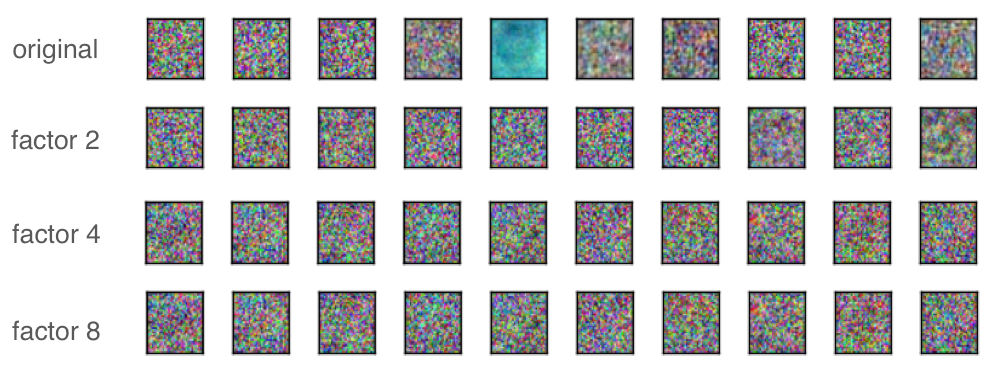}
    \caption{CIFAR100 color: 1-layer DFAE with 3100 hidden units}
    \label{fig: CIFAR100_color_features}
  \end{subfigure}
  \caption{Features learnt by a single layer DFAE on MNIST and CIFAR100 images.}
  \label{fig: 1layer_features}
\end{figure}

Next, we looked at filters or features learnt by the single layer DFAEs as shown in Figure \ref{fig: 1layer_features}. Feature detectors that correspond to the final hidden layer of the network were visualized. Each hidden neuron $y_j$ has an associated vector of weights $\mathbf{W}_{j}$ that it uses to compute a dot product with an input example. These \textit{weight vectors or filters} have the same dimensionality as the input allowing us to visualize them as images, highlighting the aspects of the input to which a hidden unit is sensitive. The goal of visualizing feature detectors was to examine qualitatively the kind of feature detectors learnt from the downsampled images and compare them to those learnt from full-resolution input.

For MNIST images, a single layer DFAE learns neuron like features when given the original input. When the input was downsampled, it was forced to learn stroke like features. A curiously similar result was observed by Vincent et al. \cite{vincent2010stacked}, where their denoising autoencoder learnt global structures when it was trained on corrupted inputs. Our DFAE also learnt increasingly global features when the input is downsampled correspondingly. However the ability to learn useful features deteriorated when the input was downsampled beyond a factor $4$. For instance, when given an input downsampled by a factor $7$, a majority of the features learnt were superimpositions of two digits and this was reflected in the images reconstructed as shown in Figure \ref{fig: MNIST_output}. On the other hand, the filters learnt on CIFAR100 images does not look meaningful. In some cases the network learnt a specific color gradients or locally circular blobs which probably enabled it to be better at reconstructing low frquency shape information and landscapes particularly well. Since we did not whiten the images, nor used image patches during training, the noise modeled by the DFAE for natural images was not surprising.

\begin{figure}[ht] 
  \begin{subfigure}[b]{0.5\linewidth}
    \centering
    \includegraphics[width=0.99\linewidth]{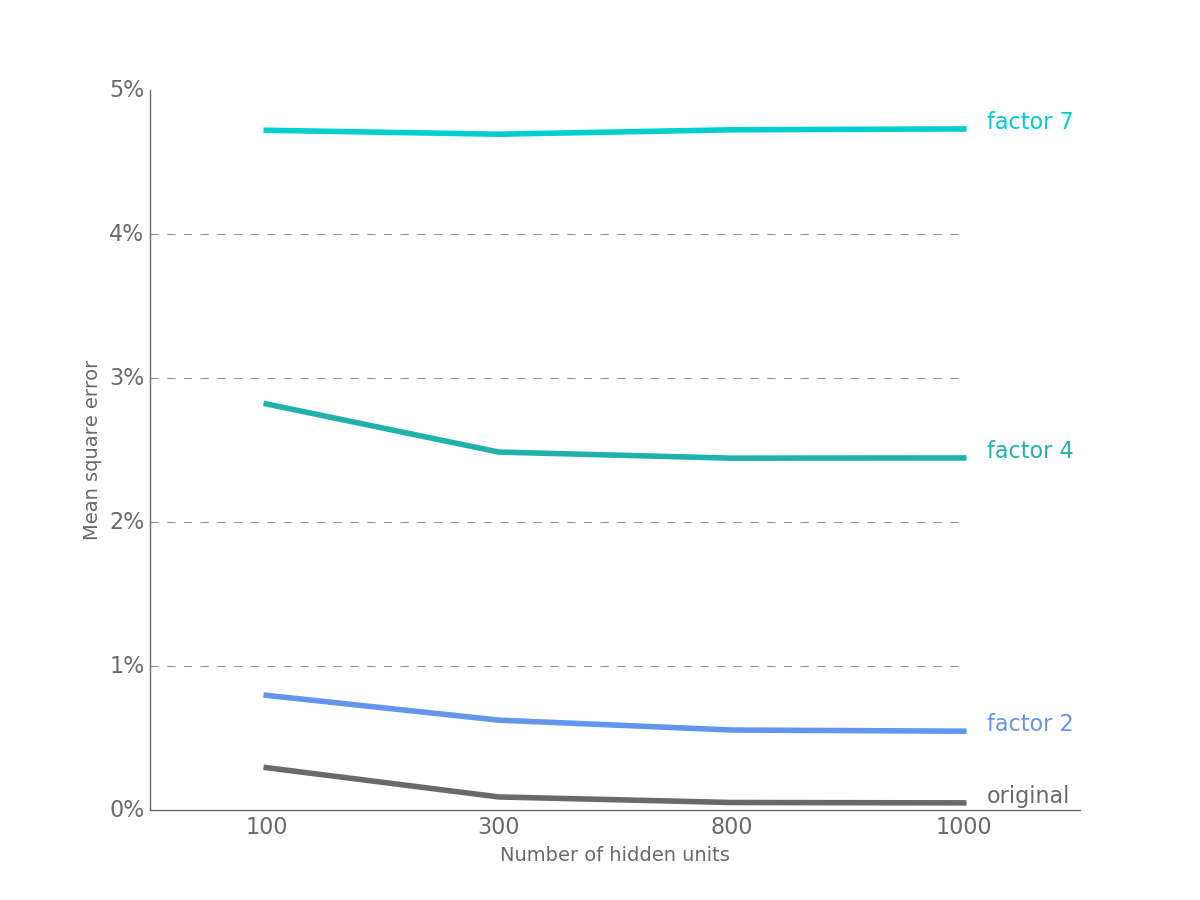} 
    \caption{MNIST: 1-layer DFAE} 
    \label{fig: M1layer_capacity} 
  \end{subfigure}
  \begin{subfigure}[b]{0.5\linewidth}
    \centering
    \includegraphics[width=0.99\linewidth]{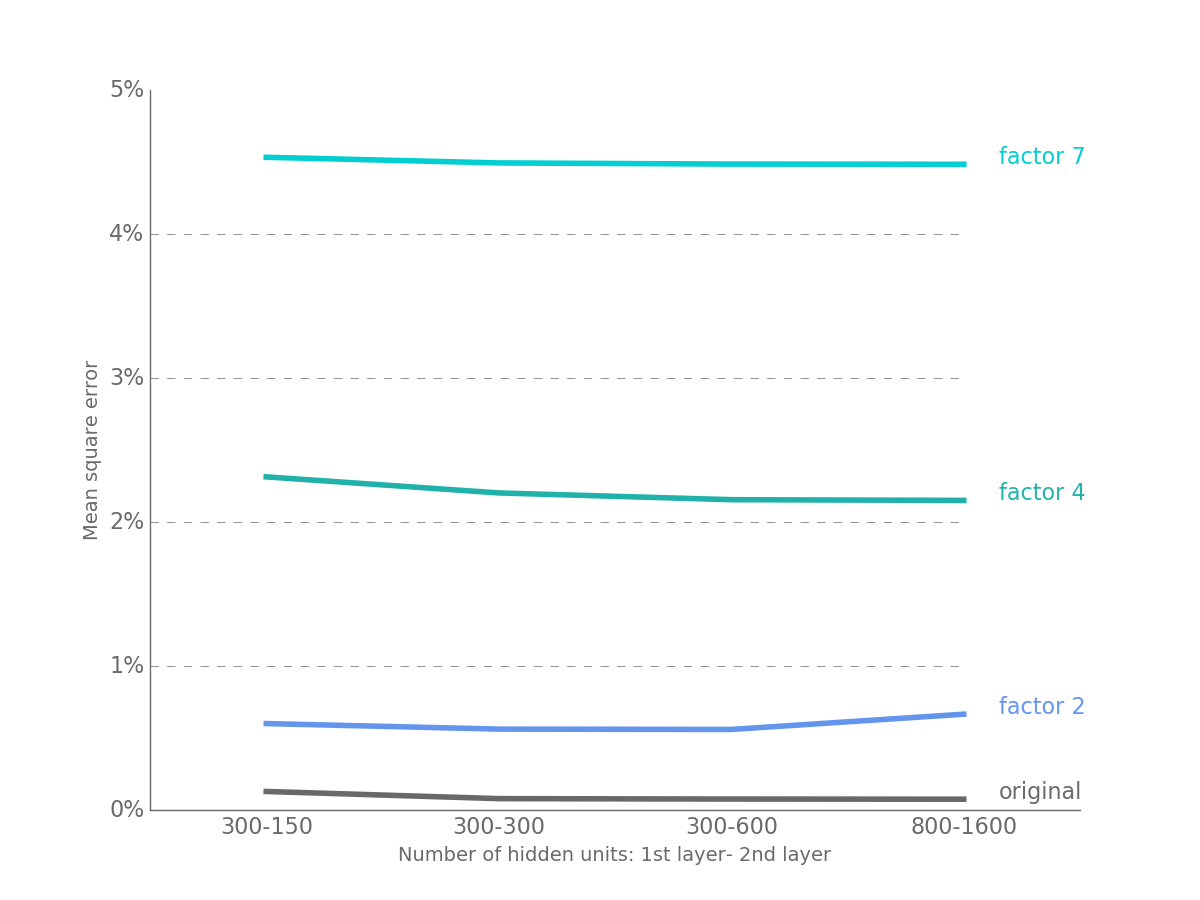} 
    \caption{MNIST: 2-layer DFAE} 
    \label{fig: M2layer_capacity} 
  \end{subfigure} 
  \begin{subfigure}[b]{0.5\linewidth}
    \centering
    \includegraphics[width=0.99\linewidth]{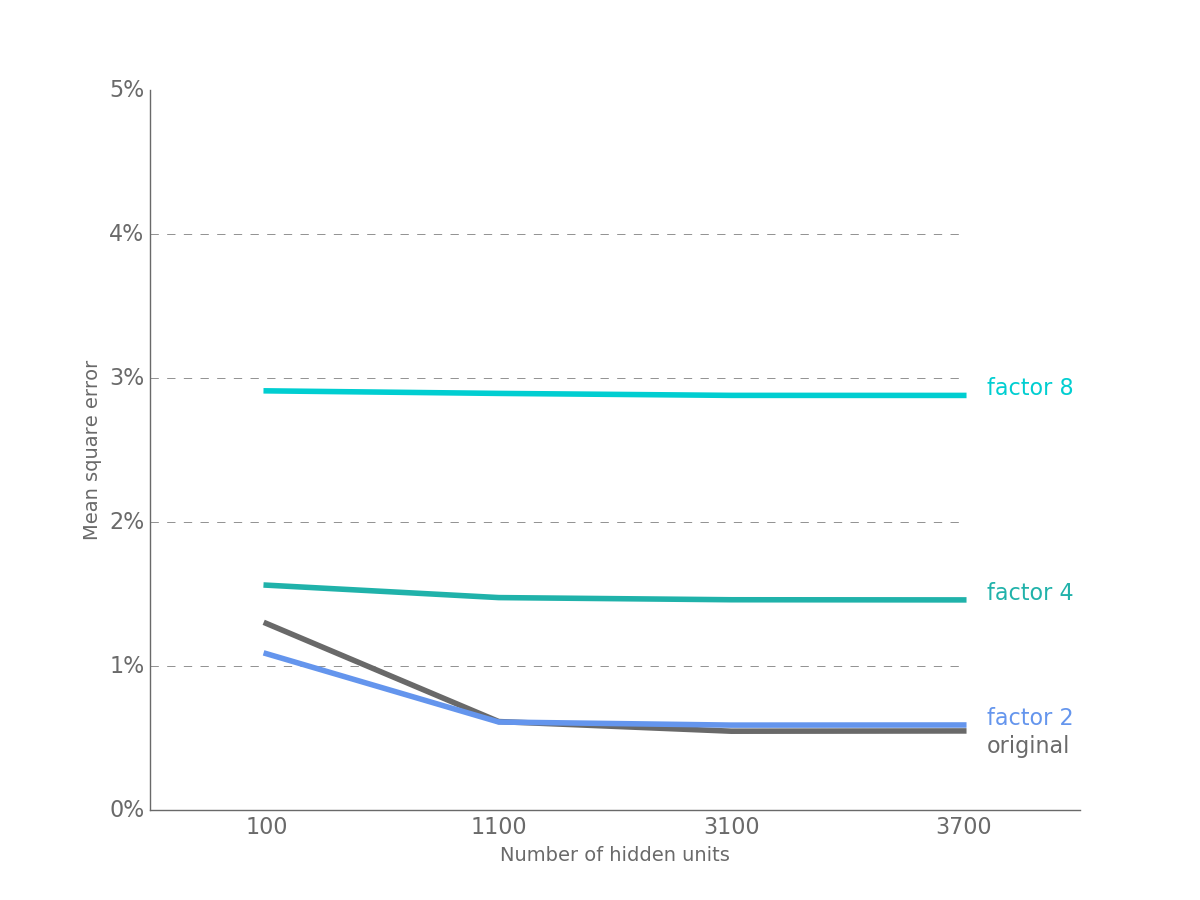} 
    \caption{CIFAR100 color: 1-layer DFAE} 
    \label{fig: CifarC_capacity} 
  \end{subfigure}
  \begin{subfigure}[b]{0.5\linewidth}
    \centering
    \includegraphics[width=0.99\linewidth]{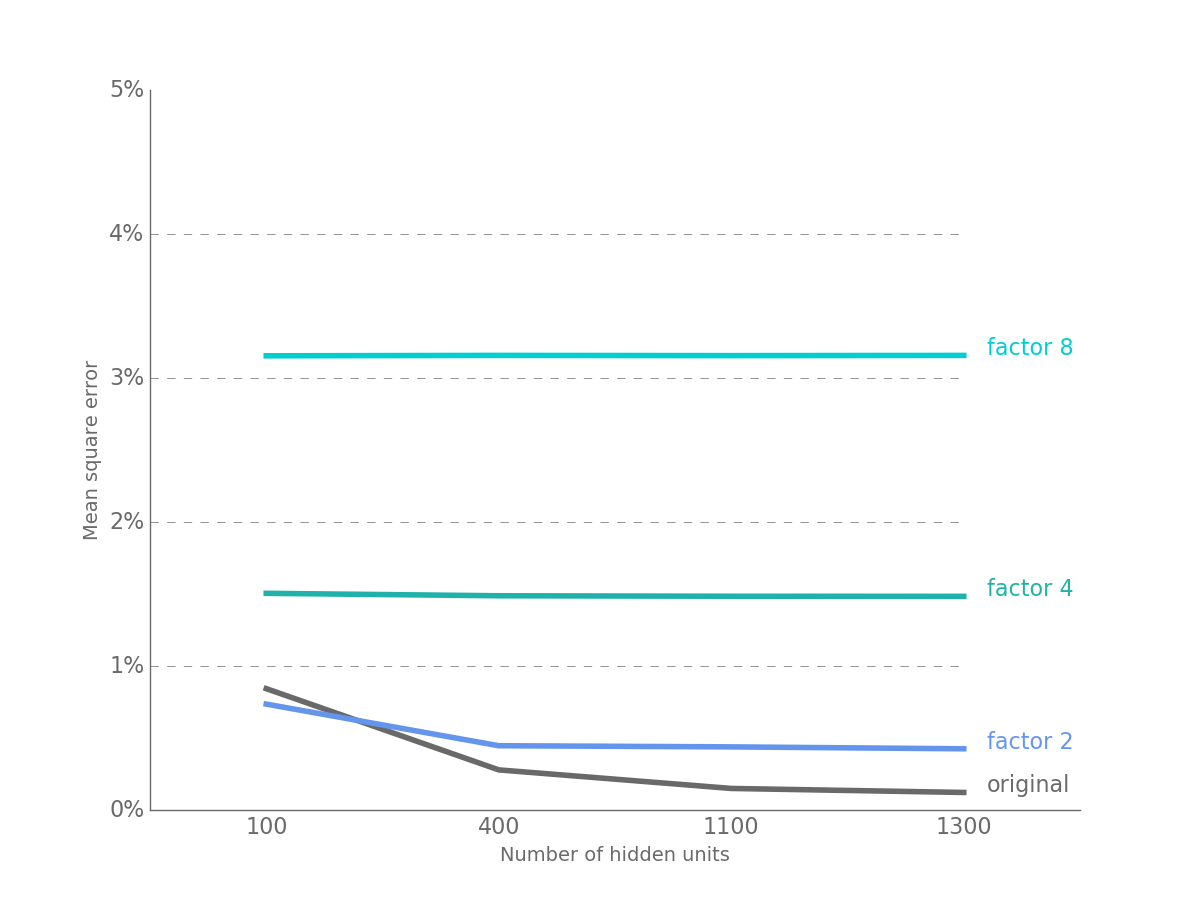} 
    \caption{CIFAR100 grayscale: 1-layer DFAE} 
    \label{fig: CifarG_capacity} 
  \end{subfigure} 
  \caption{Error rates of a DFAE when it's capacity is increased by the number of hidden units and layers. The error rates do not change when the breadth and depth of the network is creased beyond 100 units.}
  \label{fig: DFAE_capacity} 
\end{figure}

To understand how the number of hidden units and layers affect performance of DFAEs, we increased the breadth and/or depth of the DFAE. Figure \ref{fig: DFAE_capacity} show that the performance of DFAE does not improve drastically if the network is given additional capacity both in breadth (number of hidden units) and in depth (number of layers). The DFAE error rates stabilized when the number of hidden units was increased beyond 100. Note that the number of hidden units was varied according to the original input size. Therefore for 28 x 28 MNIST images the range of hidden units varied from 800 hidden units (rounded from 784 input size). Similarly, for CIFAR100 images increasing the number of hidden units of the DFAE did not improve performance of either CIFAR100 color or grayscale images. A pilot tests with networks upto 4-layers showed that performance on MNIST or CIFAR100 images did not improve significantly with increasing number of layers.

\subsection{Reconstructing foveated inputs}

Until now, we evaluated DFAEs on uniformly downsampled images but this kind of input is unrealistic from those received by the retina. In this section, we evaluate DFAEs on foveated inputs, {\bf SCT-R} and {\bf FOV-R}, as described in section $5.1$. As discussed in the introduction, the human visual system makes effective use of these kinds of foveated inputs. From a machine learning perspective, it is desirable to recognize or classify images from degraded configurations, which in turn will reduce the need for carefully pruned and preprocessed datasets during training.

\begin{figure}[ht]
  \begin{subfigure}[b]{0.45\textwidth}
  	\centering
    \includegraphics[width=\textwidth]{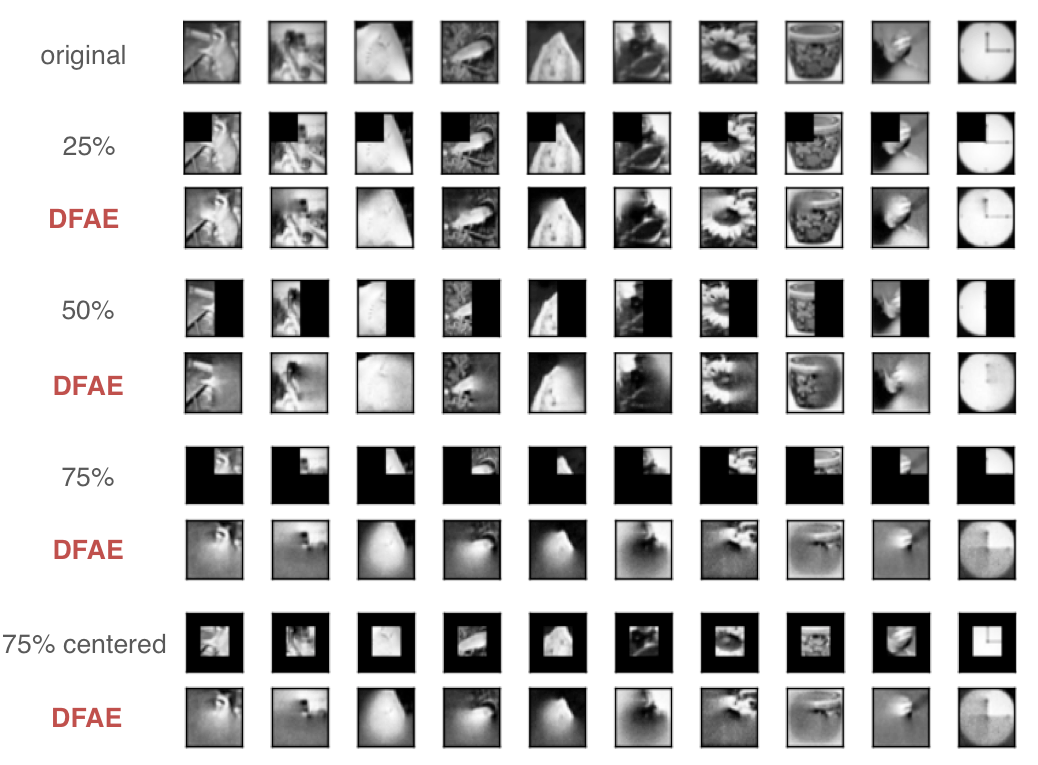}
    \caption{Scotoma,{\bf SCT-R}, reconstruction}
    \label{fig: RemovedRegions}
  \end{subfigure}
  \hfill
  \begin{subfigure}[b]{0.46\textwidth}
  	\centering
    \includegraphics[width=\textwidth]{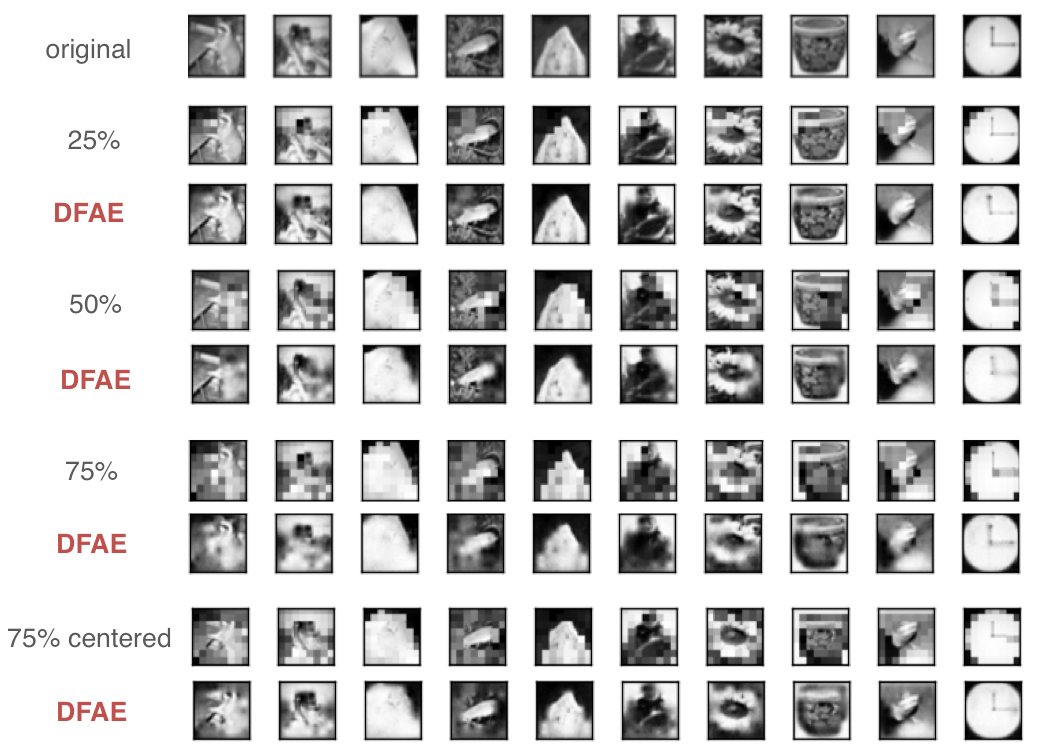}
    \caption{Foveated input, {\bf FOV-R}, reconstruction}
    \label{fig: FoveatedRegions}
  \end{subfigure}
  \begin{subfigure}[b]{\textwidth}
    \centering
    \includegraphics[width=0.7\textwidth]{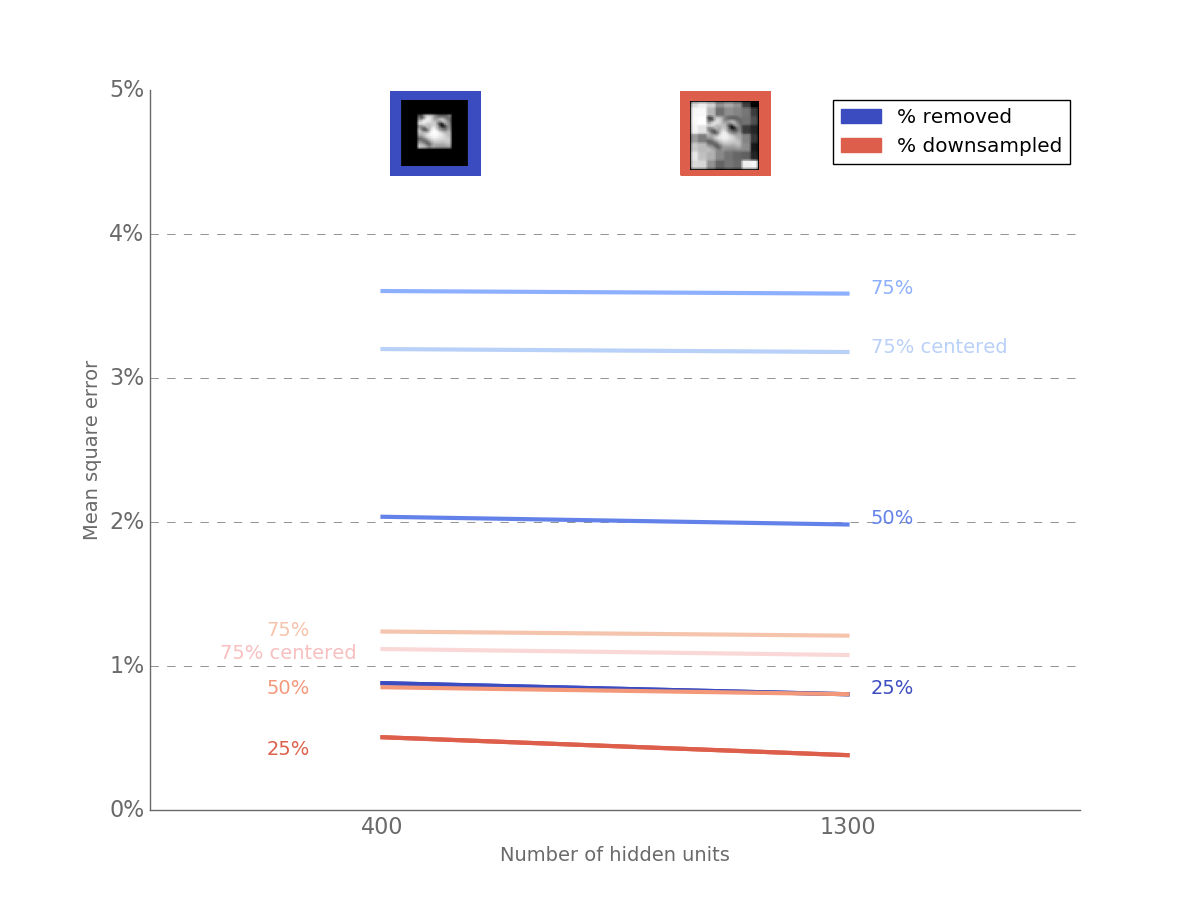}
    \caption{Error rates of 1-layer DFAE with scotoma and foveated inputs}
    \label{fig: MSE_Foveatedinputs}
  \end{subfigure}

  \caption{Reconstruction examples and errors rates of 1-layer DFAE with foveated input types}
  \label{fig: FoveatedImages}
\end{figure}

The rationale for having scotoma-like regions in the input was to test whether the available input contained enough information to reconstruct the rest of the image. The dataset used was grayscale CIFAR100 images. Variable sized areas of region ($r = $ 25\%, 50\%, 75\%, 75\% centered) were removed from the original input. The location of removal was chosen randomly from the four quadrants of the input image, except for the condition where 75\% of the image around the center was removed. Since a majority of the input images have a subject of interest, we tested if the central region contained enough information to reconstruct the rest of the image.

The reconstructions in Figure \ref{fig: RemovedRegions} show the DFAE does not perform well when $r >$ 25\%. When $r = $ 50\% of the input is removed, the DFAE can reconstruct landscapes and reconstruct shape information and symmetry, demonstrating it's ability to extract low frequency information. When $r = $ 75\% and 75\% centered, the reconstruction process breaks down and DFAE cannot predict the input beyond the given region of information. The filters learnt under these conditions look similar to the grayscale version of Figure \ref{fig: CIFAR100_color_features} with bigger smooth blobs over blacked out regions of the input.
 
 In {\bf FOV-R} inputs, $r$ is the same as {\bf SCT-R} inputs and we chose to use downsampling factor $4$ for regions outside the fovea since previous experiments revealed that DFAEs cannot reconstruct inputs downsampled beyond this factor. Figure \ref{fig: FoveatedRegions} shows the reconstructed images from {\bf FOV-R} inputs and Figure \ref{fig: MSE_Foveatedinputs} show the error rate of reconstruction. The cluster of red lines with lower error rates show that the DFAE performed considerably well with {\bf FOV-R} than {\bf SCT-R} inputs The performance was better (~1\% error for $r = $ 75\% centered) than an DFAE trained with uniformly downsampled inputs (1.5\% error). This result is not surprising, given that {\bf FOV-R} contains additional information from regions outside the fovea. These results suggests that a small number of foveations containing rich details might be all these neural networks need to extract contents of the input in higher detail. 

\subsection{Reconstructing color from foveated inputs}

It is well known that the human visual system loses chromatic sensitivity towards the periphery of the retina. Recently, there has been interest in how deep networks, specifically convolutional neural networks (CNNs), can learn to color grayscale images \cite{Dahl} and learn artistic style \cite{gatys2015neural}. Specifically in Dahl's \cite{Dahl} reconstructions from grayscale images, numerous cases of the colorized images produced were muted or sepia colored. The problem of colorization which is inherently ill-posed was treated as a classification task in these studies. Can DFAEs perceive color if it is absent in the input? 

\begin{figure}[!tbp]
  \begin{subfigure}[b]{\textwidth}
  	\centering
    \includegraphics[width=\textwidth]{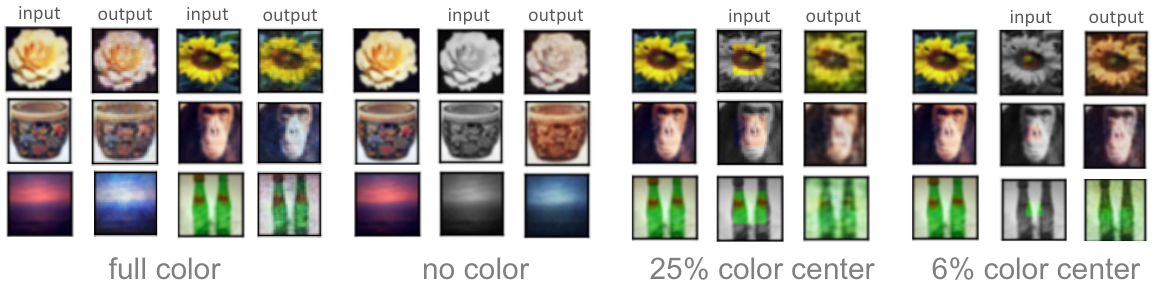}
    \caption{Color reconstruction of grayscale, {\bf ACH-R}, inputs}
    \label{fig: ACH-R}
  \end{subfigure}
  \hfill
  \begin{subfigure}[b]{\textwidth}
  	\centering
    \includegraphics[width=\textwidth]{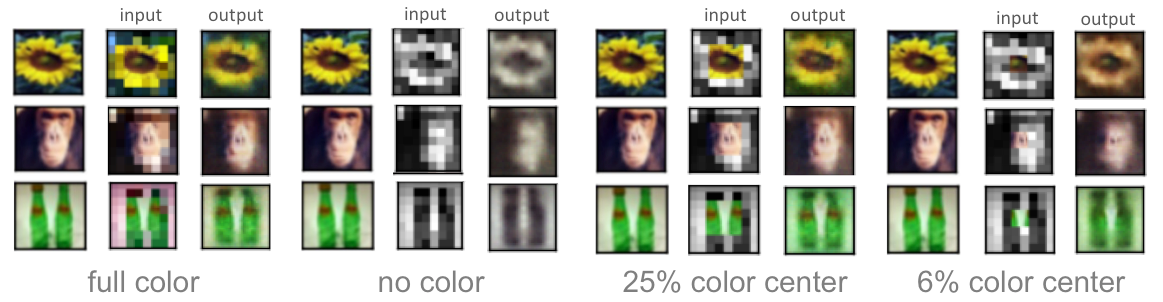}
    \caption{Color reconstruction of foveated, {\bf FOVA-R}, inputs}
    \label{fig: FOVA-R}
  \end{subfigure}
  \hfill

  \begin{subfigure}[b]{\textwidth}
    \centering
    \includegraphics[width=0.7\textwidth]{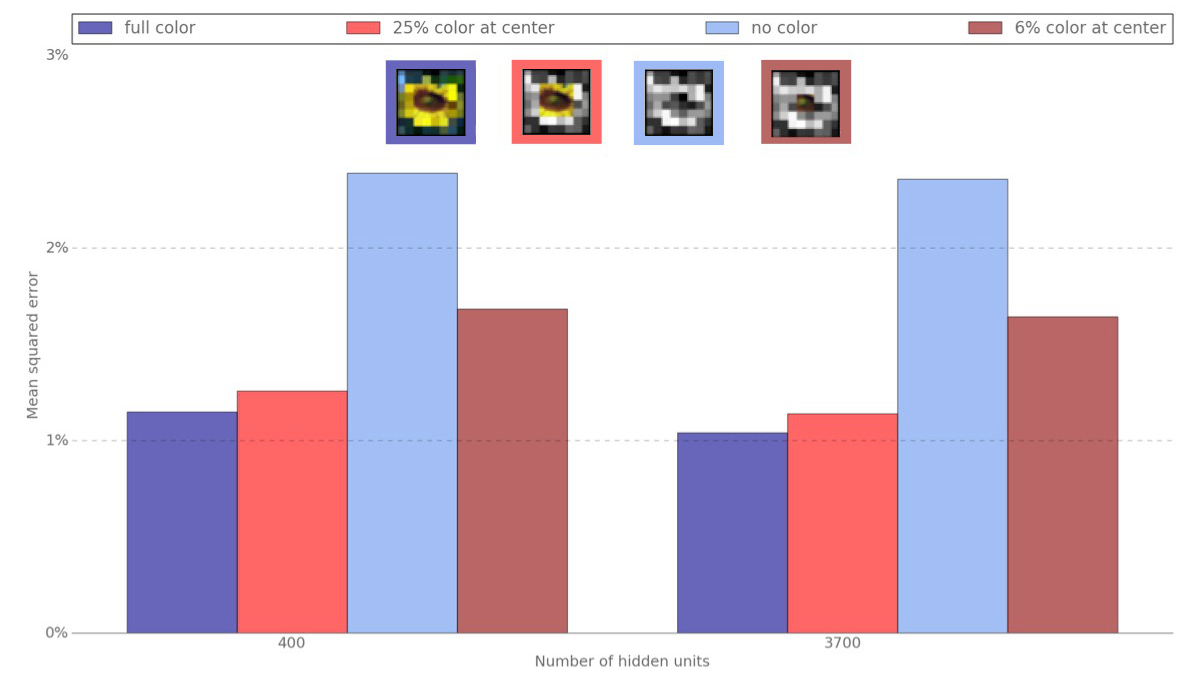}
    \caption{Color reconstruction of foveated, {\bf FOVA-R}, inputs}
    \label{fig: ColorFoveatedRegions}
  \end{subfigure}
  \begin{subfigure}[b]{\textwidth}
    \centering
    \includegraphics[width=0.7\textwidth]{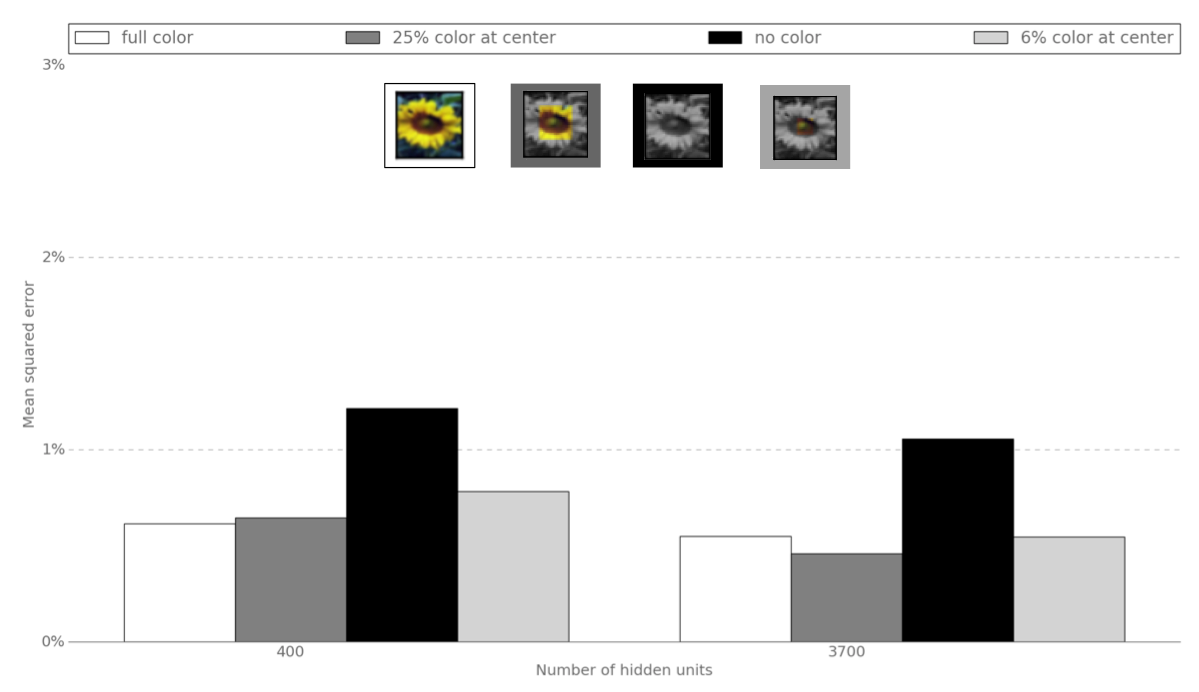}
    \caption{Color reconstruction of grayscale, {\bf ACH-R}, inputs}
    \label{fig: ColorBWinputs}
  \end{subfigure}

  \caption{Color reconstruction examples and errors rates of 1-layer DFAE}
  \label{fig: ColorReconstruction}
\end{figure}

We investigated this question using {\bf ACH-R} and {\bf FOVA-R} inputs described in section $5.1$. The regions of color tested were $r = $ 0\% or no color, 6\%, 25\% and 100\% or full color. Figure \ref{fig: ACH-R} and \ref{fig: FOVA-R} show examples of color reconstructions of the these input types. When the DFAE is trained with full color {\bf ACH-R} inputs, it can make mistakes in reconstructing the right color as seen in Figure \ref{fig: ACH-R}. For example: it colors the yellow flower as pink and the purplish-red landscape as blue. When the input is grayscale (no color, $r =$ 0\%), the colorizations are gray, muted, sepia toned or simply incorrect in the case of landscapes. But if there is a ``fovea'' of color, the single layer DFAE can reconstruct the colorizations correctly. Ofcourse, if the ``fovea'' of color is reduced, i.e. 6\%, the color reconstruction accuracy falls off but not too drastically. For example, it predicts a yellowish tone for the sunflower among a bed of brown leaves. The critical result is that the performance difference between 100\% or full colored inputs and ``foveated'' color inputs is small as seen in Figure \ref{fig: ColorFoveatedRegions} and \ref{fig: ColorBWinputs}. These results suggest that color reconstructions can be just as accurate if these networks can figure out the color of one region of the image accurately as opposed to every region in the image. Similar to the human visual system, these networks are capable of determining accurate colors in the periphery if color information is available at foveation.

\section{Conclusions}

The key finding in this paper is that current deep architectures are capable of learning useful features from low fidelity inputs. As discussed in the introduction, the human visual system uses sequential foveations to gather information about their surroundings. In each foveation, only a fraction of the input is in high resolution. We studied the capability of deep networks to learn in the face of minimal information, specifically foveated inputs. Our results indicate a single layer DFAE can reconstruct low fidelity inputs better than existing upsampling algorithms and remarkably, color reconstructions with foveated inputs are just as good with full colored inputs.

In general, our model achieves these results using a shallow network with only a small number of hidden units. We investigated how the capacity of the DFAE, in terms of layers and number of hidden units, interacts with foveated inputs. We found that small shallow networks were capable of learning good representation, especially low frequencies in the input. As noted, the performance of the the DFAE was qualitatively better on the MNIST digit images than the natural images. Firstly, the MNIST dataset contains 6000 training examples for each class compared to the CIFAR100 dataset which contains 500 training examples for each class. Secondly, the shape of the digits (a low frequency feature) is prominent in the MNIST dataset but not in the natural images which contains texture, multiple objects, contrast variation adding to the high frequency noise. The noise to signal ratio is lower in MNIST dataset in general which helped DFAE learn better representations.

Color information is obviously important to the human visual system but our results show that the performance of the DFAE does not improve significantly with color images as seen in Figures \ref{fig: CifarC_capacity} and \ref{fig: CifarG_capacity}. But color information was important in improving accuracy when the DFAE colorized images from foveated inputs.

Does an image (of scene or object) consist of a single or multiple image regions that are predictive of the contents of the image? In this paper, we focused on a single foveated region to test how that specific region was predictive of the rest of the image. Future studies can investigate which regions of an image are most predictive. How many of such regions exist within an image? Do these regions generalize across a class of images? How can we combine these regions to reconstruct the image?

In general, foveated inputs enabled the DFAE to learn the best representations overall in terms of image contents and color. This gives us hope that we can learn useful feature representations when full resolution input is not available and with a small computational budget. In future work, we want to study models that can make shifts of attention to improve the representation on demand as needed for the associated task.

\subsubsection*{}
\small{

}
\end{document}